\documentclass{article}

    \PassOptionsToPackage{numbers, sort&compress}{natbib}

 \usepackage[dandb, final]{neurips_2025}

\usepackage[utf8]{inputenc} 
\usepackage[T1]{fontenc}    
\usepackage{hyperref}       
\usepackage{url}            
\usepackage{booktabs}       
\usepackage{amsfonts}       
\usepackage{nicefrac}       
\usepackage{microtype}      
\usepackage{xcolor}         

\newcommand{\abw}[1]{{\tiny $\pm #1$}}

\usepackage{array}
\usepackage{booktabs}
\newcommand{\name}[0]{\textsc{MultiFloor3D}}
\newcommand{\datasetname}[0]{\textsc{HouseLayout3D}}

\newcommand{\ie}[0]{\emph{i.e.}}
\newcommand{\eg}[0]{\emph{e.g.}}

\definecolor{brickred}{RGB}{178, 34, 34}
\definecolor{forestgreen}{RGB}{34, 139, 34}

\newcommand{\cmark}{\textcolor{forestgreen}{\ding{51}}}
\newcommand{\xmark}{\textcolor{brickred}{\ding{55}}}

\usepackage{algorithmic}
\usepackage{algorithm2e}
\usepackage{multirow}
\usepackage{pifont}
\usepackage{float}
\usepackage{graphicx}
\usepackage{amsmath}
\usepackage{amssymb}
\usepackage[numbers,sort&compress]{natbib}

\usepackage{caption} 

\usepackage{xcolor}
\definecolor{cvprblue}{RGB}{0,113,188}
\hypersetup{
    colorlinks=true,
    linkcolor=cvprblue,
    citecolor=cvprblue,
    urlcolor=cvprblue
}
\usepackage{wrapfig}

\usepackage{titlesec}
\titlespacing{\section}{0pt}{1px}{0px}
\titlespacing{\subsection}{0pt}{1px}{0px}

\usepackage{enumitem}

\makeatletter
\renewcommand\paragraph{\@startsection{paragraph}{4}{0pt}%
  {0.25\baselineskip minus 0.1\baselineskip}
  {-0.5em}
  {\normalfont\bfseries}}
\makeatother

\title{HouseLayout3D: A Benchmark and Training-Free Baseline for 3D Layout Estimation in the Wild}

\author{%
  \vspace{-15px}\\
  \textbf{Valentin Bieri}$^{1}$ \hspace{8px}
  \textbf{Marie-Julie Rakotosaona}$^{2}$ \hspace{8px}
  \textbf{Keisuke Tateno}$^{2}$ \vspace{5px} \\
  \textbf{Francis Engelmann}$^{3}$ \hspace{8px}
  \textbf{Leonidas Guibas}$^{3}$ \vspace{5px} \\
  $^{1}$ETH Zurich \hspace{8px}
  $^{2}$Google \hspace{8px}
  $^{3}$Stanford University \\
  \vspace{-5 px}
}

\begin{document}
\maketitle

\vspace{-22pt}
\begin{center}
    \captionsetup{type=figure}
    \includegraphics[width=\textwidth]{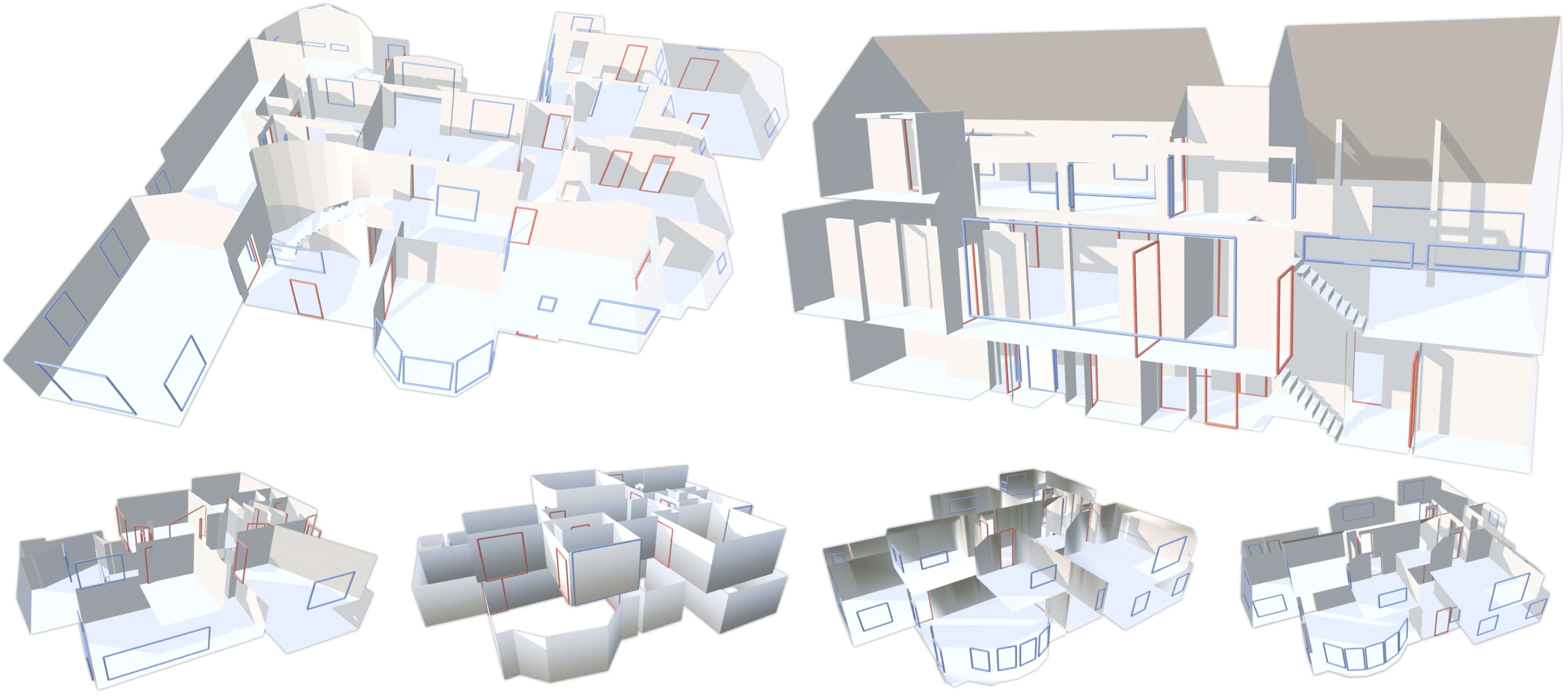}\\
    \vspace{-3px}
    \begin{small}
    \begin{tabular}{cccc}
    \hspace{3.1cm} & \hspace{3.1cm} & \hspace{3.1cm} & \hspace{2.75cm} \\
    \textit{RoomFormer}~\cite{yue2023connecting} & \textit{SceneScript}~\cite{avetisyan2024scenescript} &
    \textit{\name{}} (Ours) & \textit{Ground truth}\\
    \end{tabular}
    \end{small}
    \vspace{3px}
    \captionof{figure}{
    \emph{Top:} We present \datasetname{}, a benchmark for 3D house layout estimation with greater diversity than existing datasets, including multi floor buildings and detailed annotations for doors, windows, and staircases.
    \emph{Bottom:} We introduce \name{}, a training free approach for 3D layout estimation that improving over existing methods on both our benchmark and prior datasets.} 
    \label{fig:teaser}
\end{center}

\begin{abstract}
Current 3D layout estimation models are primarily trained on synthetic datasets containing simple single room or single floor environments. As a consequence, they cannot natively handle large multi floor buildings and require scenes to be split into individual floors before processing, which removes global spatial context that is essential for reasoning about structures such as staircases that connect multiple levels.
In this work, we introduce \datasetname{}, a real world benchmark designed to support progress toward full building scale layout estimation, including multiple floors and architecturally intricate spaces. We also present \name{}, a simple training free baseline that leverages recent scene understanding methods and already outperforms existing 3D layout estimation models on both our benchmark and prior datasets, highlighting the need for further research in this direction. Data and code are available at: \url{https://houselayout3d.github.io}.
\end{abstract}
\section{Introduction}
Having spatial awareness of the surrounding 3D layout is a key requirement for many perception algorithms \cite{yue2023connecting, ccelen2025housetour, zheng2025wildgs, jin2024multiway} and robotic systems \cite{lemke2024spotcompose, zurbrugg2024icgnet}. The goal is to derive a concise vectorized layout by abstracting a 3D scene into a set of polygons that represent structural elements such as walls, floors, ceilings, staircases, doors, and windows, while filtering out furniture and other occluders commonly found in indoor environments.

Recent state-of-the-art models for layout prediction \cite{yue2023connecting, avetisyan2024scenescript, chen2023polydiffuse} are feed-forward deep-learning models trained on large-scale synthetic datasets \cite{Structured3D, avetisyan2024scenescript} and demonstrate impressive results even on real-world scenes.
A key limitation of current models is their reliance on synthetic training data, which predominantly contains single rooms or small apartments. Such data is appealing because it can be generated automatically at scale~\cite{avetisyan2020scenecad} or created in controlled settings~\cite{Structured3D}, yet it lacks the complexity of large real buildings. As a result, models trained on these datasets struggle to generalize to buildings with many rooms and cannot handle multi-floor layouts.
One workaround is to partition a building into individual floors or rooms, process each part independently, and then merge the results. However, this removes global context that is important for local reasoning, for example when identifying staircases that connect multiple floors, and it requires post hoc integration to support building-level tasks such as localization~\cite{miao2024scenegraphloc, wuest2025unloc} or scene-level reasoning~\cite{zhang2025open}.

To drive progress in 3D layout prediction for large-scale multi-floor buildings, we introduce \datasetname{}, a benchmark built on real-world scans from Matterport3D~\cite{Matterport3D}. The dataset contains architecturally complex buildings spanning as many as five floors and up to forty rooms per floor, including diverse room types and partially open spaces that remain difficult for current room-centric methods. We provide detailed manual annotations of structural elements such as walls, floors, ceilings, staircases, windows, and doors, including the opening direction of each door.

Inspired by recent progress in reconstruction and segmentation, we propose \name{}, a training free approach for large scale 3D layout estimation. By combining modern 3D scene reconstruction with a layout fitting strategy, we show that a simple method can outperform existing approaches on the more challenging task of layout estimation in multi floor buildings.

Our experiments on \datasetname{} reveal clear limitations of current state of the art methods in complex multi floor buildings. In contrast, our approach produces more accurate and coherent layouts, particularly in challenging multi level structures. We hope that these findings, together with our benchmark, will encourage further research in scalable multi floor 3D layout estimation.

In summary, our contributions are:
\begin{itemize}[left=0pt, itemsep=0pt, topsep=0pt]
\item We introduce \datasetname{}, the first benchmark for 3D layout estimation in large scale multi floor buildings.
\item We present \name{}, a training free baseline leveraging modern reconstruction and segmentation models, achieving improved performance over existing deep learning approaches.
\item Through extensive experiments we expose the challenges faced by current layout estimation methods, motivating progress in this problem setting.
\end{itemize}

\section{Related Work}

\paragraph{Manhattan Scene Layout.}

Early approaches to layout estimation commonly assume a Manhattan world and solve a constrained optimization problem using detected walls, as in Scan2Bim~\cite{scan2bim}, or using detected corners, as in DuLaNet~\cite{yang2019dulanetdualprojectionnetworkestimating}, LayoutNet~\cite{zou2018layoutnet}, and FloorNet~\cite{liu2018floornetunifiedframeworkfloorplan}.
A notable exception by Ochmann \emph{et al.}~\cite{OCHMANN2019251} relaxes the Manhattan constraint by subdividing the 3D space into cells and formulating layout estimation as an integer linear program, allowing angled walls.

\paragraph{2D Scene Layout.}
Early methods such as~\cite{cabral2014piecewiseplanarshortestpath} infer 2D floorplans by computing shortest paths around free space, while Floor-SP~\cite{chen2019floorspinversecadfloorplans} extends this idea with a room segmentation network. RoomFormer~\cite{yue2023connecting} estimates semantic floorplans with a transformer. HovSG~\cite{werby23hovsg} combines BEV point-density maps with 2D object detection to construct a scene graph of floors, rooms, and objects, but without recovering their 3D geometry. This class of approaches remains fundamentally limited to 2D predictions.

\paragraph{3D Scene Layout.}
End-to-end learning has driven recent progress in 3D layout estimation: SceneCAD~\cite{avetisyan2020scenecad} predicts layouts and object boxes using a graph network, and SceneScript~\cite{avetisyan2024scenescript} introduces a structured scene language for joint prediction of walls, openings, and objects.
Compact scene abstraction has also emerged as a complementary direction, for example SuperDec~\cite{fedele2025superdec}, which decomposes indoor environments into superquadric primitives and highlights the value of structured, geometry-efficient representations.
However, most available training data consists of single rooms~\cite{avetisyan2020scenecad} or simple one-floor layouts~\cite{Structured3D,avetisyan2024scenescript,ZInD}, often synthetic~\cite{Structured3D,avetisyan2024scenescript}, unfurnished~\cite{ZInD}, or restricted to Manhattan geometries~\cite{matterportLayout3D}. Other datasets only capture partial scenes~\cite{rozumnyi2023estimatinggeneric3droom,matterportLayout2D}. This lack of diverse building-scale data limits generalization and leaves current models ill suited for complex multi-floor environments.

\section{The \datasetname{} Dataset}

\begin{wrapfigure}[18]{r}{0.5\textwidth}
    \vspace{-23pt}
    \centering
    \vspace{-1em}
    \includegraphics[width=\linewidth]{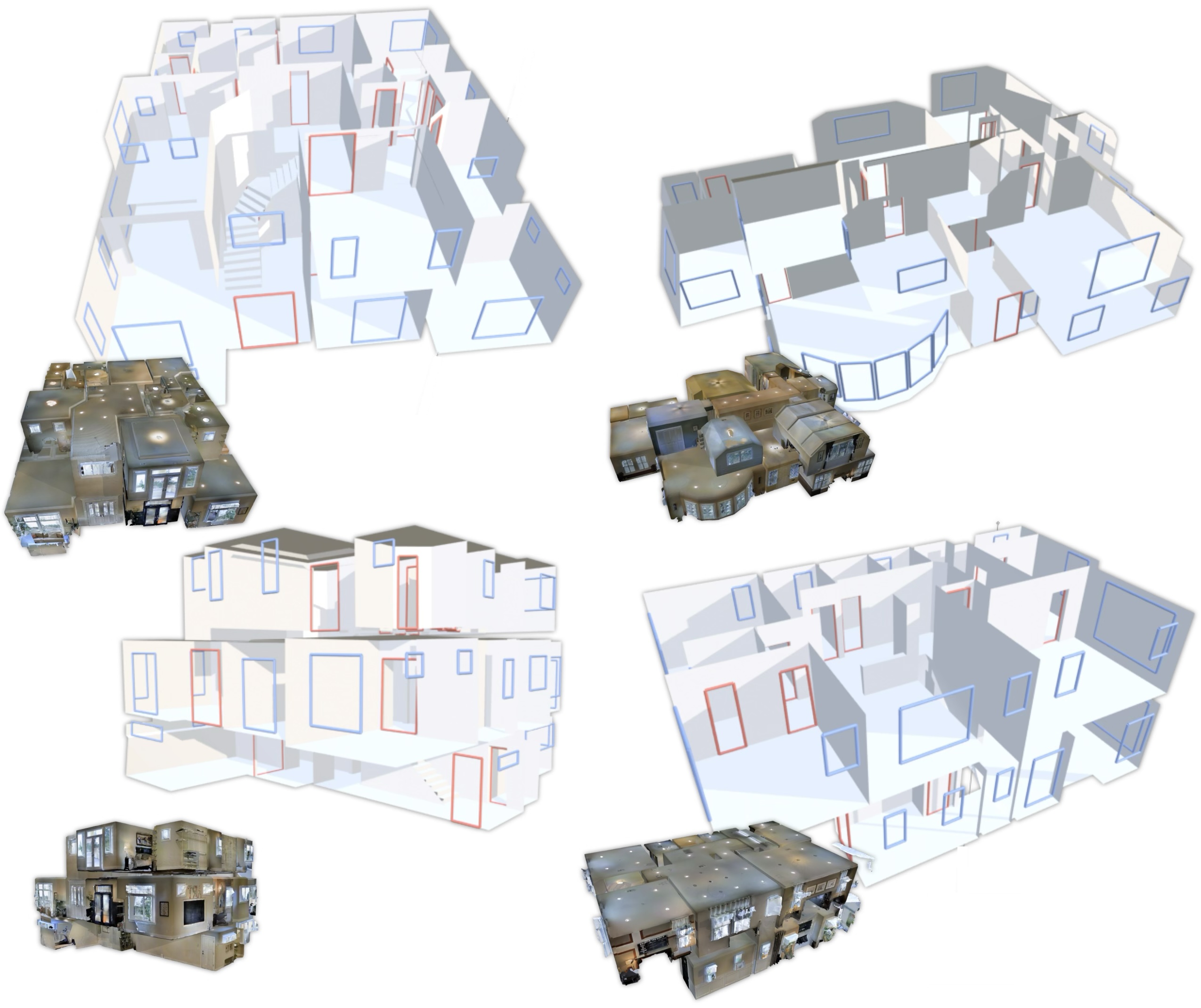}
    \caption{\textbf{Examples of our \datasetname{}.}
    Our dataset includes multi-floor houses with annotations for walls, floors, ceilings and stairs, as well as windows {\color{blue} \emph{(blue)}} and doors {\color{red} \emph{(red)}}.
    We also show the corresponding 3D meshes from MP3D~\cite{Matterport3D}.}
    \label{fig:houselayout3d_dataset}
\end{wrapfigure}

We introduce a new dataset of hand-annotated CAD layouts derived from the Matterport3D~\cite{Matterport3D} (MP3D) dataset. Example annotations are shown in Fig.~\ref{fig:houselayout3d_dataset}. Unlike prior works \cite{avetisyan2020scenecad, avetisyan2024scenescript}, this is the first real-world dataset to provide CAD annotations for large-scale, multi-floor houses, encompassing numerous rooms, staircases, windows, and doors. Each structural element is annotated as a polygon in 3D space. Since our dataset is annotated on 3D meshes from MP3D \cite{Matterport3D}, it inherits their per-vertex room ids and object instances. 

\paragraph{Dataset Statistics.}
The dataset includes 16 buildings, 33 distinct levels, and 317 rooms, captured across more than 26,000 RGB-D frames. Its scale is comparable to the validation split of ScanNet~\cite{dai2017scannet}. In total, we annotated 292 doors, 379 windows, and 34 staircases. The lower number of doors compared to rooms is due to many spaces, such as hallways and dining areas, being connected by open passages or staircases rather than actual doors. Each building comprises between 1 and 5 levels and contains between 4 and 40 rooms.
The annotation time varies depending on the building's size and the number of rooms, typically ranging from 4 to 10 hours per building. All annotations undergo visual verification by separate expert annotators.
Tab.~\ref{tab:dataset_comparisons} compares and summarizes properties across different datasets.

\begin{table}[t]
\centering
\setlength{\tabcolsep}{3px}
\resizebox{\textwidth}{!}{%
\begin{tabular}{l c c c c c c c c}
\toprule
\rotatebox{0}{Dataset} &
\rotatebox{0}{Real-world} &
\rotatebox{0}{Multi-room} &
\rotatebox{0}{Multi-floor} &
\rotatebox{0}{Full Scenes} &
\rotatebox{0}{Windows, Doors} &
\rotatebox{0}{Objects} &
\rotatebox{0}{Depth} &
\rotatebox{0}{3D Layouts} \\
\midrule
SceneCAD~\cite{avetisyan2020scenecad}                   & \cmark & \textcolor{orange}{(\ding{51})} & \xmark   & \cmark & \cmark & \cmark & \cmark & \cmark \\
ASE~\cite{avetisyan2024scenescript}                       & \xmark   & \cmark & \xmark   & \cmark & \cmark & \cmark & \cmark & \cmark \\
Stru3D~\cite{Structured3D}                     & \xmark   & \cmark  & \textcolor{orange}{(\ding{51})} & \cmark & \cmark & \cmark & \cmark  & \cmark \\
Zillow Indoor~\cite{ZInD}              & \cmark & \cmark  & \textcolor{orange}{(\ding{51})} & \cmark & \xmark  & \xmark  & \xmark & \xmark \\
MP3D-Layout~\cite{matterportLayout2D}                & \cmark & \xmark   & \xmark   & \xmark  & \cmark & \cmark & \cmark & \cmark \\
Zou et al~\cite{matterportLayout3D}                & \cmark & \xmark   & \xmark   & \xmark  & \xmark  & \xmark & \cmark & \cmark \\
CADEstate~\cite{rozumnyi2023estimatinggeneric3droom}                & \cmark & \cmark  & \xmark  & \xmark & \textcolor{orange}{(\ding{51})}   & \xmark & \xmark & \cmark \\
FloorNet~\cite{liu2018floornetunifiedframeworkfloorplan}                   & \cmark & \cmark  & \cmark  & \cmark & \textcolor{orange}{(\ding{51})} & \xmark & \xmark & \xmark \\
\datasetname{} (Ours) & \cmark & \cmark & \cmark & \cmark & \cmark & \cmark & \cmark & \cmark \\
\bottomrule
\end{tabular}
}
\caption{\textbf{Dataset Comparisons} of existing dataset benchmarks for evaluating 3D layouts estimation.}
\label{tab:dataset_comparisons}
\end{table}

\paragraph{Annotation Tool and Labeling Details.}
To annotate the 3D scans, we use a free academic license of Scasa's PinPoint~\cite{scasa2025pinpoint}, a specialized software for building modeling from point clouds. It enables precise 3D geometry extraction even in occluded or incomplete areas through intuitive tools that automatically snap to edges and corners, streamlining the annotation process. In the 3D scans, doors are typically open, so we annotate both the current open position and the expected closed position, along with the opening direction. For doors that appear closed in the scans, we infer the opening direction from the door hinge locations in the RGB images.
For window annotations, we utilize the existing annotations from MP3D~\cite{Matterport3D}, projecting them onto the nearest annotated wall plane and fitting axis-aligned rectangles.

\section{Method}

Given $N$ input RGB images of a scene, our goal is to recover a simple 3D layout represented by polygons. Each polygon is assigned a label from a fixed set of classes: wall, floor, ceiling, stairs, door, or window. The resulting layout is organized into a scene graph whose nodes correspond to rooms and whose edges correspond to doors or stairs, with each polygon associated either with a room node or with an edge in this graph.

Fig.~\ref{fig:pipeline} shows an overview of our four-stage approach. First, we reconstruct a 3D mesh of the scene. Second, we extract the main structural elements (floors, walls, ceilings) to form a \textit{skeleton} of the layout. Third, we fit a layout \textit{prototype} to this skeleton using geometric and semantic cues. Finally, we parse the prototype into a scene graph from which we derive the final layout.

\subsection{Mesh Reconstruction from RGB Images} 
\label{sec:mesh_generation}
Given a set of unposed 2D images, we follow DN-Splatter~\cite{turkulainen2024dnsplatter} to obtain a triangle mesh and depth maps for each frame. DN-Splatter uses COLMAP~\cite{schoenberger2016sfm} poses together with a 2D depth model to train a 3D Gaussian Splatting~\cite{kerbl3Dgaussians} reconstruction, and then produces a Poisson surface~\cite{kazhdan2006poisson} by sampling from the depths rendered by 3DGS. In our implementation, we use the Metric3D~\cite{yin2023metric3dzeroshotmetric3d} depth model.

\subsection{Layout Skeleton Extraction from Mesh}
\label{sec:skeleton_extraction}
Once the mesh is obtained, we extract a minimal and reliable geometry that serves as the layout \textit{skeleton} using a pre-trained 2D segmentation model. The skeleton should contain only the geometry intended for the final layout. To separate such geometry, we distinguish four semantic categories:

\begin{itemize}[left=0pt, itemsep=0pt, topsep=0pt]
\item \textbf{Structural Components} (walls, ceilings, floors, and large furniture such as closets): these form the core of the layout skeleton and provide accurate geometry that should be preserved.
\item \textbf{Geometrically Inaccurate Surfaces} (windows, mirrors): these often suffer from poor depth estimates and are therefore excluded from the skeleton.
\item \textbf{Objects} (small furniture and household items): these are removed from the skeleton but later help infer missing layout regions.
\item \textbf{Stairs}: due to their complexity and diversity, they are detected and processed separately.
\end{itemize}

\begin{figure*}[t]
    \centering
    \includegraphics[width=\linewidth]{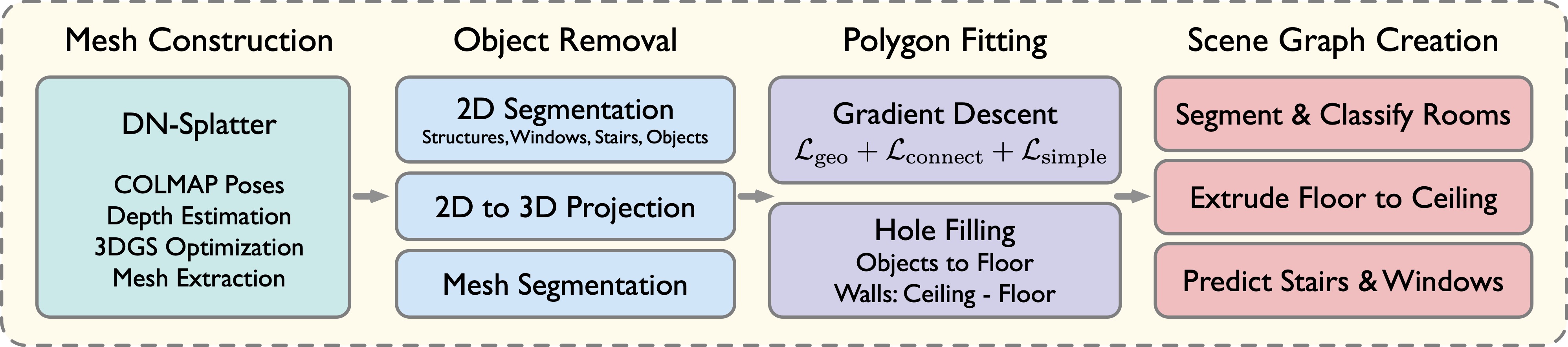}
    \includegraphics[width=\linewidth]{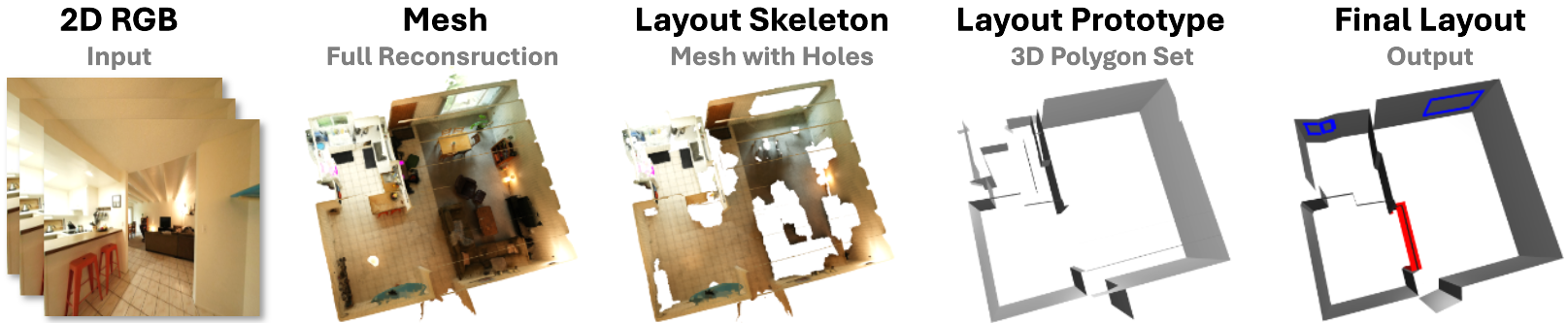}
    \caption{\textbf{Illustration of the \name{} model for 3D layout estimation.} }
    \label{fig:pipeline}
\end{figure*}

To construct the skeleton, we first segment the 3D mesh into the four semantic categories. We run the OneFormer model~\cite{jain2023oneformer} on the input images and map its output classes~\cite{lin2015microsoftcococommonobjects} to our categories. To transfer these labels to the mesh, we back-project M = 5000 randomly sampled pixels per image along with their predicted class into 3D, assigning each back-projected point to the nearest mesh vertex and accumulating class votes.
We then refine the segmentation by clustering the mesh into \textit{superpoints} following the preprocessing of~\cite{robert2024scalable} and assigning each vertex the majority label within its cluster. This yields a mesh labeled with our semantic classes.
From this mesh, we extract the \textit{layout skeleton} by selecting the structural components, and isolate the \textit{object} and \textit{stair} subsets for later processing.
See Fig.~\ref{fig:pipeline} \emph{(bottom)} for an illustration of a layout skeleton.

\subsection{Fitting a Layout Prototype to the Skeleton}
\label{sec:prototype_fitting}
We observe significant artifacts in the layout skeletons, including holes and unobserved regions. For example, areas hidden behind furniture, or areas corresponding to windows are missing.
In this stage, we use geometric and semantic information to correct the artifacts and infer a more complete \textit{layout prototype}.
To this end, we run an optimization that aims to improve the completeness of the obtained skeleton: We first initialize a collection of planar 3D polygons $\mathcal{P}$  from the layout skeleton. In particular, each segmented superpoint (see Sec.~\ref{sec:skeleton_extraction}) of the skeleton is fitted to one or more planes. We then optimize the \textit{vertex positions} and \textit{plane equations} of the polygons using three main objectives:
\begin{itemize}[left=0pt, itemsep=0pt, topsep=0pt]
    \item $\mathcal{L}_{\text{geo}}$ reconstructs an accurate scene geometry
    \item $\mathcal{L}_{\text{connect}}$ produces a continuous and connected geometry
    \item $\mathcal{L}_{\text{simple}}$ produces a mesh with low vertex count
\end{itemize}
During the optimization, we constrain the vertices of each polygon to be coplanar. We also allow and encourage polygons to share vertices. The initialization and the implementation of vertex constraints and shared vertices is detailed in the supplementary material.

\paragraph{Definitions.} Given a polygon $P$ consisting of edges $E$ and a point $p \in \mathbb{R}^3$ we define the point-to-polygon distance $D_{pp}(P, p)$ as the minimal distance between $p$ and any point on the surface of $P$. For $e\in E$ we define the point-to-edge distance $D_{pe}(p, e)$ as the minimal distance between $p$ and any point on the line segment representing $e$.

\paragraph{Losses.} We fit the polygon set $\mathcal{P}$ using gradient descent and three losses.
The first loss $\mathcal{L}_{\text{geo}}$ encourages the polygons to reconstruct the  original geometry and respect the \textit{observed empty space}:
\begin{equation}
\label{eq:L_geom}
\mathcal{L}_{\text{geom}} = \mathcal{L}_{\text{prox}} + \mathcal{L}_{\text{empty}}
\end{equation}
$\mathcal{L}_{\text{prox}}$ penalizes the distance of each vertex $v \in V_{\text{skeleton}}$ of the Layout Skeleton to the closest polygon surface:
\begin{equation}
\label{eq:L_proximity}
\mathcal{L}_{\text{prox}} = \sum_{v \in V_{\text{skeleton}}} \min_{P \in \mathcal{P}} D_{pp}\bigl(v, P\bigr)
\end{equation}
To prevent occluding the observed empty space (\ie{},  the space we believe to be empty based on the depth maps), we sample a set $L$ of line segments using the input camera poses and computed depth maps. Each line segment extends from the camera pose to the back-projected depth. We then penalize line segment-polygon intersections as follows: If a line segment $l$ intersects a polygon, the nearest polygon edge $e^*$ should be moved closer to the intersection point $p_{inter}$.
\begin{equation}
\label{eq:L_empty}
\mathcal{L}_{\text{empty}} = \sum_{l \in L} \; \sum_{\substack{P \in \mathcal{P} \\ l \cap P \neq \varnothing \\ D_{pe}( p_{inter},\, e^* \bigr) \leq \tau_{\text{inter}}}} 
D_{pe}\bigl( p_{inter},\, e^*)
\end{equation}
\[
\text{where } p_{inter} = l \cap P \quad \text{and} \quad e^* = \operatorname*{argmin}_{e' \in \text{edges}(P)} D_{pe}\bigl(v, e'\bigr).
\]
Note that we ignore intersections with $D_{pe}( p_{inter},\, e^* )$ greater than the threshold $\tau_{\text{inter}}$ to avoid noise from intersections far from the polygon boundary.

The second loss $\mathcal{L}_{\text{connect}}$ prevents small gaps and encourages shared boundaries by making polygons attract vertices. Concretely, $\mathcal{L}_{\text{connect}}$ penalizes the distance from each polygon vertex to the closest surface of another polygon: 
\begin{equation}
\label{eq:L_connect}
\mathcal{L}_{\text{connect}} = \sum_{P \in \mathcal{P}} \sum_{v \in \mathrm{vertices}(P)} \min_{P' \in \mathcal{P}, ~~P' \neq P} D_{pp}\bigl(v, P'\bigr)
\end{equation}
As for $\mathcal{L}_{\text{empty}}$, we ignore points with $ D_{pp}\bigl(v, P'\bigr)$ greater than a threshold.

The third loss encourages simplicity and smooth polygon boundaries. $\mathcal{L}_{\text{simple}}$ penalizes the length of all edges that are not shared by at least two polygons. (\ie{}, not all edge vertices are shared). Intuitively, $\mathcal{L}_{\text{simple}}$ promotes shared edges (for instance, an edge between two walls) to represent the scene while edges that are not shared are shrunk until they are eliminated.
\begin{equation}
\label{eq:L_simple}
\mathcal{L}_{\text{simple}} = \sum_{P \in \mathcal{P}} \sum_{e \in \operatorname{edges}(P)} \mathbf{1}_{\left[ \nexists\, P' \in \mathcal{P} \setminus \{P\} : ~ e \subset P' \right]} \|e\|_2
\end{equation}
Our final loss is given by $\mathcal{L} = \mathcal{L}_{\text{geom}} + \mathcal{L}_{\text{connect}} + \mathcal{L}_{\text{simple}}$.

\paragraph{Vertex Merging}
$\mathcal{L}_{\text{simple}}$ itself does not reduce the number of vertices or polygons in the polygon set.
Instead, we periodically manually simplify $P$ by (1) merging vertex pairs with distance below $\tau_{\text{merge}}$, (2) applying the RDP~\cite{douglas1973algorithms} algorithm with tolerance $\tau_{\text{merge}}$ to the polygons individually, and (3) merging close polygons with similar normal. (Close in terms of minimal $D_{pp}$ distance among the vertices.) 
For (3) we additionally verify that the merged polygon does not increase $\mathcal{L}_{\text{prox}}$ too strongly. 
Note that step (1) is the source of shared vertices between polygons.

\paragraph{Closing Holes in the Floor.} 
We observe that there is a floor under most objects in a room. We exploit this information by projecting objects to the floor, \ie{}, we project each triangle of the \textit{object} mesh extracted in Sec.~\ref{sec:skeleton_extraction} to the plane equation of the nearest floor-classified polygon whose centroid lies below the triangle. We recompute the floor polygon from the union of the original floor polygon and the projected triangle surfaces. 

\paragraph{Closing Wall Holes.}
We extend walls to span from ceiling to floor. Specifically, we identify polygon edges of wall-classified polygons whose normals face downward, and count how many line segments in $L$ (representing observed empty space) intersect the region between each edge and the floor. If the number of intersections per $\text{cm}^2$ falls below $\tau_{\text{extend}}$, we extend that edge to the floor. We apply the same procedure to ceilings and to wall edges whose normals face upward.  
The output of this stage is the \textit{layout prototype}.

\subsection{Scene Graphs from a Layout Prototype}
\label{sec:scene_graph_parsing}

Next, we convert the prototype (a set of semantically labeled polygons) into the final layout. The layout is represented as a scene graph whose nodes correspond to rooms and whose edges correspond to doors and stairs. Each room node contains one floor along with its associated wall, ceiling, and window polygons. To obtain this structure, we first generate 2D floorplans and then extrude them into 3D. The indirection through 2D is motivated by the fact that the 3D layout prototype does not provide a clear indoor–outdoor separation and does not guarantee that the layout is closed or connected.

\paragraph{Creation of a Scene Graph of 2D Floorplans}
\label{2D-scene-graph}
In this step we use the layout prototype and its semantics to (1) identify the different levels (floors) of the building, (2) create a 2D layout (floorplan) of each level, and (3) segment each level into rooms, extracting a per-level 2D scene graph from each floor and (4) detect stairs to connect the individual levels. In the following, we provide an outline of the applied algorithms, which are detailed in the appendix.

\begin{itemize}[left=0pt, itemsep=0pt, topsep=0pt]
    \item To identify building floors, we use the floor-classified polygons of the layout prototype, merging close levels with similar heights.

    \item To create a 2D floorplan of each level, we merge each level's floor polygon(s) with suitable ceiling polygons — since ceilings are rarely occluded by objects and thus are more robustly represented in the layout prototype.

    \item To segment each level into rooms, we apply Hov-SG~\cite{werby23hovsg}'s room segmentation algorithm on each 2D floorplan (and the walls of the layout prototype). The segmentation outputs a scene graph with rooms as nodes, and \textit{openings} as edges. We consider an opening edge a \textit{door} if its width is below $1.5\,\mathrm{m}$. Otherwise, we retain its edge but label it as \textit{opening}. Furthermore, each room is associated with a room type (kitchen, office, \dots).

    \item To identify stairs, we cluster connected components of the stair mesh extracted in Sec.~\ref{sec:skeleton_extraction}. For each component, we add an edge to the scene graph between the rooms and floors it connects.
\end{itemize}

\begin{figure}[]
    \centering
    \includegraphics[width=0.75\linewidth]{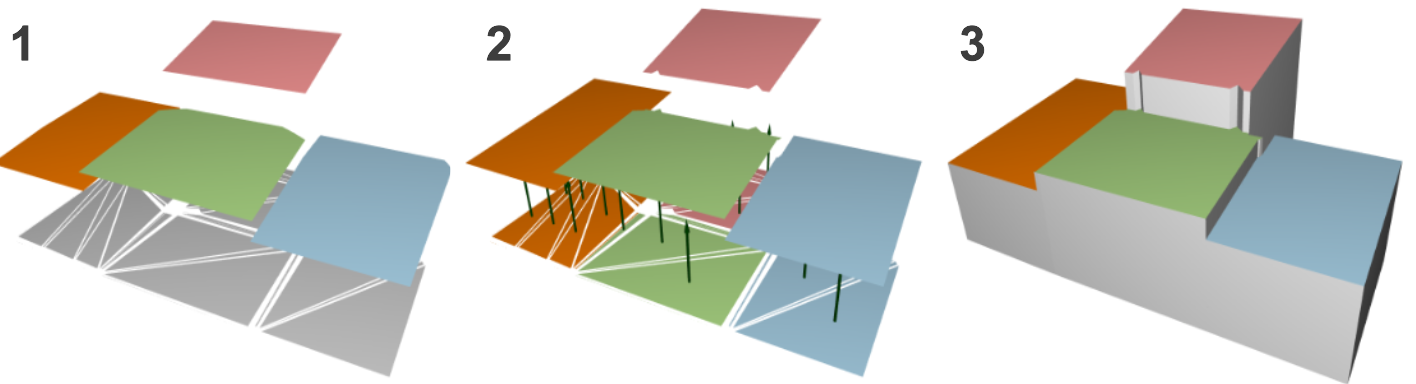}
    \caption{\textbf{Our proposed floor extrusion algorithm.}
    1) Floor triangulation.
    2) Triangles assigned to ceilings using midpoints.
    3) Triangles extruded to ceiling planes.}
    \label{fig:extrusion-graph}
\end{figure}

\paragraph{Back to 3D: Room Extrusion}
Sec.~\ref{2D-scene-graph} describes how we use the layout prototype to generate a scene graph of rooms. In this section, we propose a simple algorithm inspired by layout annotation tools~\cite{scasa2025pinpoint}, that extrudes each node's 2D floorplan to the ceiling.
For a single room, the extrusion algorithm creates a closed room shell using a 2D floorplan and a set of potential 3D ceiling polygons that at least partially cover the floorplan.
Fig.~\ref{fig:extrusion-graph} visualizes the extrusion process. Its core idea is to (1) triangulate the 2D floorplan, (2) assign each triangle to a ceiling polygon and (3) extrude each triangle to its ceiling.
Specifically, we triangulate the room's 2D floorplan using a 2D Constrained Delaunay Triangulation~\cite{CGAL:ConstrainedDelaunayTriangulation2} built from the boundary of the floorplan, the ceiling candidates' edges, and the projections of the pairwise intersection lines of the ceiling candidates' planes.
For each triangle center, we cast a ray upward. If the ray hits a ceiling candidate, we assign the triangle to that ceiling's plane.
Intuitively, this assignment partitions the floorplan by `rendering' ceiling polygons on the floor.
Triangles that do not hit a ceiling are assigned to the lowest ceiling plane reachable in the graph of unassigned triangles. (Lowest in terms of the triangle midpoint's projected z-coordinate on the target plane.)
Lastly, we extrude each floor triangle to its assigned ceiling plane. That is, we produce ceiling and floor triangles on the ceiling and floor planes respectively, and add axis-aligned wall rectangles for triangle edges coinciding with a wall in the 2D floorplan. To ensure a closed room shell, we further add vertical rectangles along potential discontinuous edges in the extruded ceiling surface. To limit complexity, we only consider the $30$ largest ceilings per room.
Details on how we add doors and stairs after extruding are provided in the appendix.

\paragraph{Window Detection}
To detect windows, we back-project the 2D window segmentation of the input images obtained in Sec.~\ref{sec:skeleton_extraction} onto layout walls and cluster the result. 
Concretely, we create rays for window-classified pixels and intersect them with the walls of our 3D layout. We then filter outliers~\cite{localoutlierfactor}, split the 
points by wall instance, and run DBSCAN~\cite{dbscan} for each wall to identify window clusters. To every cluster with at least $k=10$ vertices, we fit an axis-aligned bounding rectangle. Finally, we predict a window for every rectangle with height and width greater than $30$ cm.

\begin{table*}[t]
\centering
\setlength{\tabcolsep}{2pt}
\resizebox{\textwidth}{!}{%
\begin{tabular}{l *{2}{c} *{2}{c} *{2}{c} *{2}{c} | *{3}{c}}
\toprule
 & \multicolumn{2}{c}{\textbf{Structures}} & \multicolumn{2}{c}{\textbf{Doors}} & \multicolumn{2}{c}{\textbf{Windows}} & \multicolumn{2}{c}{\textbf{Stairs}} & \multicolumn{3}{c}{\textbf{Depth}} \\
\cmidrule(lr){2-3} \cmidrule(lr){4-5} \cmidrule(lr){6-7} \cmidrule(lr){8-9} \cmidrule(lr){10-12}
\textbf{Method} & {F1@0.5} & {Avg F1} & {F1@0.5} & {Avg F1} & {F1@0.5} & {Avg F1} & {F1@0.5} & {Avg F1} 
& \(\Delta_5\) & \(\Delta_{10}\) & {\#Vertices} \\
\midrule
RoomFormer~\cite{yue2023connecting} {\footnotesize (per floor)}  &
$0.24$\abw{0.06} &
$0.22$\abw{0.06} &
$0.23$\abw{0.10} &
$0.20$\abw{0.09} &
$0.07$\abw{0.06} &
$0.07$\abw{0.04} & -- & -- &
$24.9$\abw{11.5} &
$32.9$\abw{14.9} &
$764.9$\\
RoomFormer~\cite{yue2023connecting} {\footnotesize (per room)} &
$0.18$\abw{0.14} &
$0.16$\abw{0.12} &
$0.18$\abw{0.14} &
$0.16$\abw{0.12} &
$0.08$\abw{0.08} &
$0.09$\abw{0.07} & -- & -- &
$37.3$\abw{10.4} &
$44.8$\abw{10.7} &
$1134.5$\\
SceneScript~\cite{avetisyan2024scenescript} {\footnotesize (per floor)} &
$0.28$\abw{0.11} &
$0.26$\abw{0.08} &
$0.23$\abw{0.26} &
$0.20$\abw{0.23} &
$0.16$\abw{0.18} &
$0.15$\abw{0.17} & -- & -- &
$22.5$\abw{8.6} &
$33.8$\abw{11.7} &
$\mathbf{677.1}$\\
SceneScript~\cite{avetisyan2024scenescript} {\footnotesize (per room)} &
$0.23$\abw{0.12} &
$0.21$\abw{0.11} &
$0.31$\abw{0.26} &
$0.28$\abw{0.23} &
$0.11$\abw{0.11} &
$0.10$\abw{0.09} & -- & -- &
$23.5$\abw{7.2} &
$32.9$\abw{6.7} &
$1333.6$\\
\name{} (Ours) &
$\mathbf{0.40}$\abw{0.10} & $\mathbf{0.38}$\abw{0.10} &
$\mathbf{0.55}$\abw{0.16} & $\mathbf{0.44}$\abw{0.15} &
$\mathbf{0.43}$\abw{0.29} & $\mathbf{0.38}$\abw{0.22} &
$\mathbf{0.42}$\abw{0.48} & $\mathbf{0.41}$\abw{0.44} &
$\mathbf{61.1}$\abw{9.2} & $\mathbf{76.3}$\abw{7.9} & $1957.0$
\\
\bottomrule
\end{tabular}
}
\caption{\textbf{Scores on \datasetname{}.} Performance comparison with state-of-the-art layout estimation methods in terms of average and standard deviation across scenes. Structures include wall, floor and ceilings. \name{} is the only method predicting stairs.}
\label{tab:quantitative_our_dataset}
\end{table*}

\section{Experiments}

\begin{figure*}
\vspace{-10pt}
    \centering
    \includegraphics[width=\linewidth]{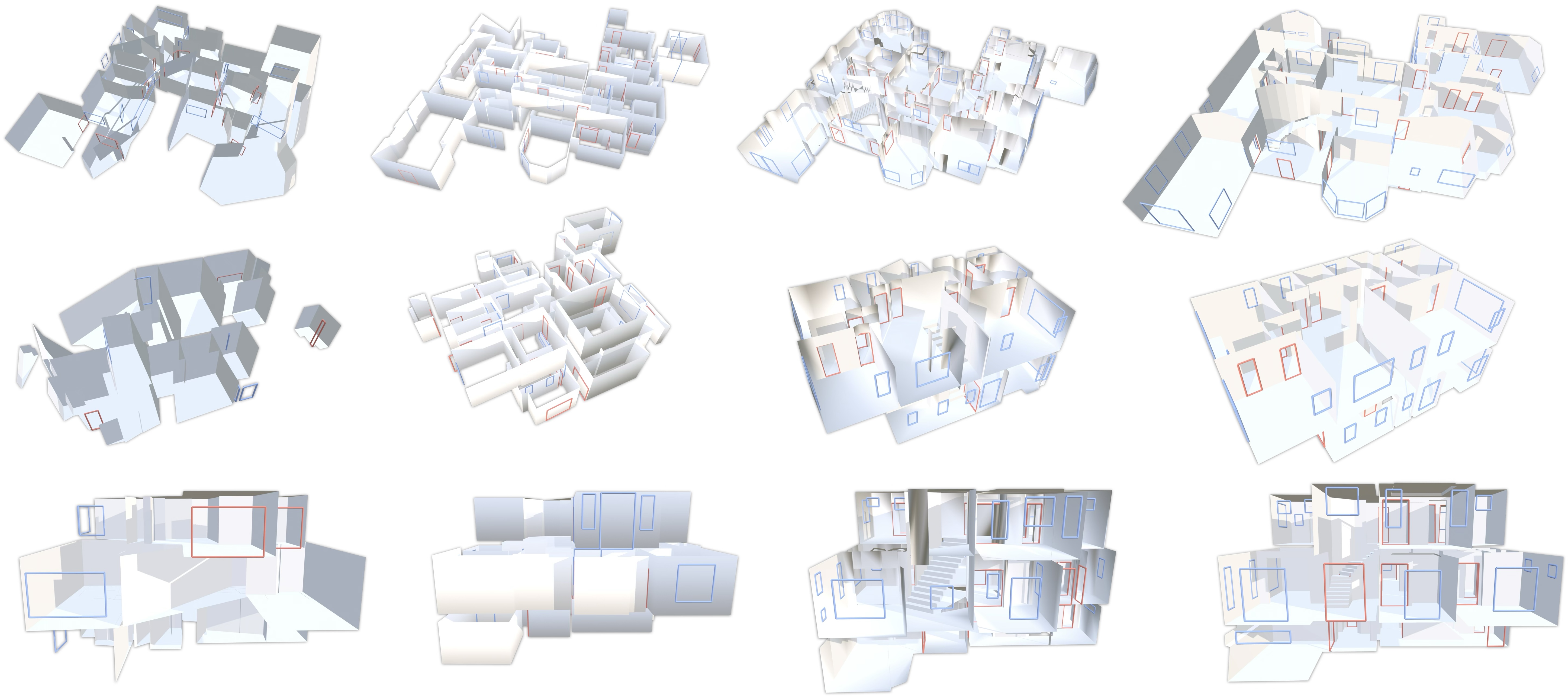}
    \begin{small}
\begin{tabular}{cccc}
\emph{RoomFormer~\cite{yue2023connecting}} & \emph{SceneScript}~\cite{avetisyan2024scenescript} &\emph{\name{} (Ours)}& \textit{Ground-truth} \\
\hspace{0.23\linewidth} & \hspace{0.23\linewidth} & \hspace{0.23\linewidth} & \hspace{0.23\linewidth} \\
\end{tabular}\end{small}
\vspace{-10px}
    \caption{\textbf{Qualitative Results on \datasetname{}.} We present layout estimation samples from our model alongside state-of-the-art methods. To enhance visualization, we apply back-face culling to the  layout meshes, allowing a clear view inside the buildings. Since SceneScript represents walls as boxes, back-face culling is ineffective; instead, we remove the added floors and ceilings for better visibility.}
    \label{fig:quali}
\end{figure*}

In this section, we introduce the metrics used to evaluate 3D room layout estimation and compare our approach to recent state of the art methods on \datasetname{} (Sec.\ref{sec:our_dataset}) and on ScanNet++\cite{scannetpp} (Sec.\ref{sec:scannetpp}). We then analyze the contribution of individual pipeline components (Sec.\ref{sec:analysis}), and conclude with qualitative results and potential applications (Sec.~\ref{sec:qualitative}).

\paragraph{Methods in Comparison.}
We compare our approach to two recent scene layout estimation methods: RoomFormer~\cite{yue2023connecting} and SceneScript~\cite{avetisyan2024scenescript}. Training these baselines on our multi floor dataset is non trivial: RoomFormer targets 2D floorplan prediction, and SceneScript is limited to four corner primitives, so we evaluate them using their publicly available weights on the full \datasetname{} dataset. Both baselines are trained on large synthetic datasets (about 100k samples), whereas our method is training free. Following~\cite{avetisyan2024scenescript}, we extrude RoomFormer’s 2D layouts into 3D. As neither baseline predicts floors or ceilings, we append floor and ceiling polygons to each predicted room to enable a fair depth evaluation.

\paragraph{Layout Metrics.}
To assess the accuracy of the estimated layouts, we adopt the F1 score based on the \textit{entity distance} $d_E$, following SceneScript~\cite{avetisyan2024scenescript}.
This metric measures the alignment between ground truth entities $E$ and predicted entities $E'$.
For rectangular entities (\eg{}, doors and windows), $d_E$ is computed as the maximum distance between corresponding corners of two rectangles of the same class:
$
d_E(E, E') = \max \Bigl\{ \| c_i - c'_{\pi(i)} \| : i = 1, \ldots, 4 \Bigr\}
$
where $\pi(i)$ denotes the optimal corner permutation obtained via Hungarian matching.
The F1 score \textit{@}$\tau$ is then computed by applying a threshold $\tau$ to $d_E$ as in ~\cite{avetisyan2024scenescript}. 
For non-rectangular entities, we introduce a generalized entity distance $d_{H}$ which allows comparison between entities with different numbers of corners.
We define $d_{H}$ as the Hausdorff distance between two polygon surfaces (\ie entities) $P$, $P'$ and their vertices $V$, $V'$:
\begin{equation}
\label{eq:entity_hausdorff}
d_{H}(P, P') = \max\Bigl\{ \max_{v \in V} D_{pp}(v,P'), \; \max_{v' \in V'} D_{pp}(v',P) \Bigr\}
\end{equation}
for the point-to-polygon distance $D_{pp}$ defined in Sec~\ref{sec:prototype_fitting}.
We then use $d_{H}$ analogously to $d_E$ to compute the F1 score for walls, floors, and ceilings.

\paragraph{Depth Metrics.}
Following \cite{matterportLayout3D}, we use input camera poses to render depth maps for the ground truth geometry $D_{GT}$ and predict layouts $D_{\text{pred}}$.
When explicit layout annotations are unavailable (\eg{}, ScanNet++~\cite{scannetpp}), depth consistency serves as a proxy for evaluating layout accuracy.
Specifically, we compute the percentage of predicted pixel depths that fall within a threshold $\tau$ cm of the GT depth:
\begin{equation}
\label{eq:layout_depth}
\Delta_{\tau} = \frac{1}{N} \sum_{i=1}^{N} \mathbf{1}_{\left[\left|D_{\text{pred}}(i) - D_{GT}(i)\right| \le \tau\right]}
\end{equation}
This $\Delta_{\tau}$ metric was introduced ~\cite{delta_1_metric} and is commonly used in monocular depth estimation~\cite{yin2023metric3dzeroshotmetric3d}.

\paragraph{Results on \datasetname{}}
\label{sec:our_dataset}
Tab.~\ref{tab:quantitative_our_dataset} shows scores for the F1-based metrics across semantic classes, and depth metrics on our \datasetname{} dataset.
For this experiment, we use the camera poses, RGB-D images and mesh of MP3D ~\cite{Matterport3D}.
As neither baseline is designed for multi-floor layout prediction, we apply them separately per floor (or per room) and then merge the per-floor (or per-room) predictions. We use the ground-truth MP3D floor and room segmentation and report scores per-floor and per-room. Our \name{} does not have access to this privileged information.

\name{} significantly outperforms state-of-the-art layout estimation methods, despite not using ground-truth floor or room segmentation. While both baselines perform better on individual rooms than full floors, this gap is smaller for SceneScript, which favors compactness (\ie{}, fewer vertices) at the cost of geometric accuracy.

\begin{table*}[tb]
\centering
\begin{minipage}[t]{0.48\textwidth}
    \centering
    \setlength{\tabcolsep}{3pt}
    \begin{small}
    \begin{tabular}{l c c c}
    \toprule
    \textbf{Method} & \textbf{\#Vertices} & \(\Delta_{5}\) & \(\Delta_{10}\) \\
    \midrule
    \textcolor{gray}{DN-Splatter Mesh}~\cite{turkulainen2024dnsplatter} & \textcolor{gray}{$354k$} & \textcolor{gray}{$84.1$} & \textcolor{gray}{$92.6$} \\
    \midrule
    RoomFormer~\cite{yue2023connecting} & $\mathbf{32.5}$ & $36.8$ & $48.9$ \\
    SceneScript~\cite{avetisyan2024scenescript} & $41.2$ & $55.1$ & $68.5$ \\
    \name{} (Ours) & $83.1$ & $\mathbf{67.8}$ & $\mathbf{84.7}$ \\
    \bottomrule
    \end{tabular}
    \end{small}
    \caption{\textbf{Scores on ScanNet++~\cite{scannetpp}.}
    Metrics evaluate depth accuracy as an approximation of layout estimation error. Scores are averaged over validation scenes.}
    \label{tab:scannetpp_results}
\end{minipage}
\hfill
\begin{minipage}[t]{0.48\textwidth}
    \centering
    \setlength{\tabcolsep}{3pt}
    \resizebox{\textwidth}{!}{%
    \begin{tabular}{l c c c}
    \toprule
    \textbf{Method} & \textbf{Avg F1} & \textbf{\#Vertices} & \textbf{Sem.} \\
    \midrule
    Input Mesh + QSlim~\cite{garland1997quadricedgedecimation}         & $0.109$ & $2000.0$ & \xmark \\
    Layout Skeleton + QSlim~\cite{garland1997quadricedgedecimation}   & $0.223$ & $2000.0$ & \xmark \\
    Layout Prototype & $0.373$ & $2553.0$ & \xmark \\
    \name{} (Ours) & $\mathbf{0.381}$ & $\mathbf{1957.1}$ & \cmark \\
    \quad (w/o prototype fitting) & $0.214$ & $2269.8$ & \cmark \\
    \quad (w/o room segmentation) & $0.359$ & $2442.2$ & \textcolor{orange}{(\ding{51})} \\
    \bottomrule
    \end{tabular}
    }
    \caption{\textbf{Ablation Study on \datasetname{}.}}
    \label{tab:pipeline_ablation}
\end{minipage}
\end{table*}

\paragraph{Results on Scannet++}
\label{sec:scannetpp}
Tab.~\ref{tab:scannetpp_results} shows additional results on the Scannet++~\cite{scannetpp} \textit{DSLR} validation split, consisting of 50 scenes captured with a monocular hand-held camera and COLMAP-generated image poses. As ScanNet++ does not provide ground truth layout annotation, we only report depth metrics as an approximation of the layout error.
Since ScanNet++ scenes are populated with objects, we use ground truth semantic annotations to ignore those points during the evaluation, as well as points on windows which are typically not well reconstructed in the ground truth laser scan.
As input for all methods, use the mesh provided by the Gaussian Splatting approach DN-Splatter~\cite{turkulainen2024dnsplatter} in the first stage of our method (Sec.~\ref{sec:mesh_generation}).
The results indicate that \name{} outperforms the baselines at the cost of compactness (larger number of vertices).

\newpage
\begin{wrapfigure}{r}{0.6\textwidth}
    \centering
    \includegraphics[width=\linewidth]{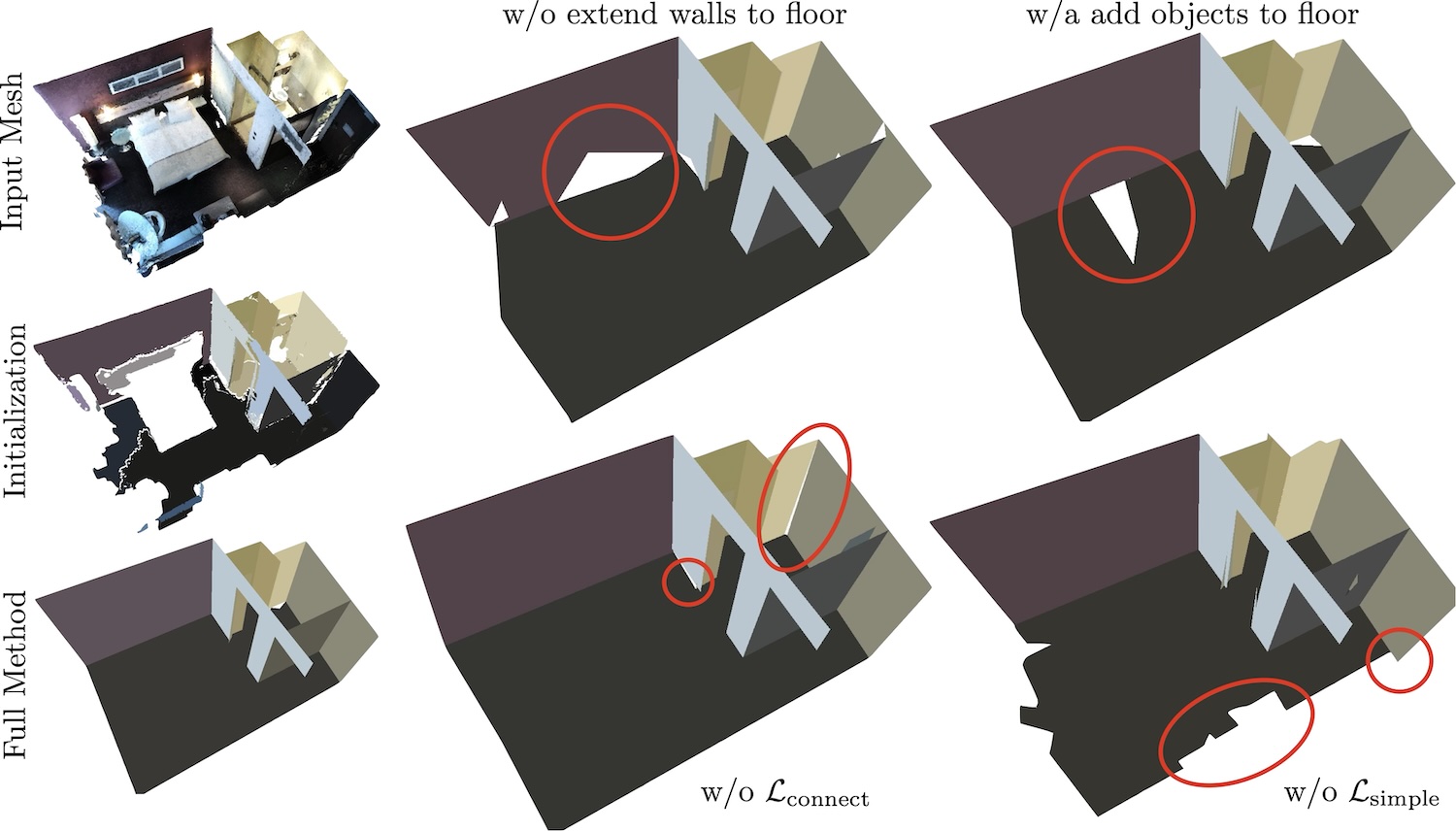}
    \caption{\textbf{Effect of Loss Terms.} \emph{Left:} input and output of our approach.
    \emph{Right:} result when ablating losses and components.
    Omitting object projection (top center) or wall extension (top right) produces holes in the layout. Without $\mathcal{L}_{\text{simple}}$, the polygon boundaries show dents. Without $\mathcal{L}_{\text{connect}}$, we observe gaps between polygons that otherwise share edges.}
    \label{fig:loss-ablation}
    \vspace{3pt}
    \centering
    \includegraphics[width=0.9\linewidth]{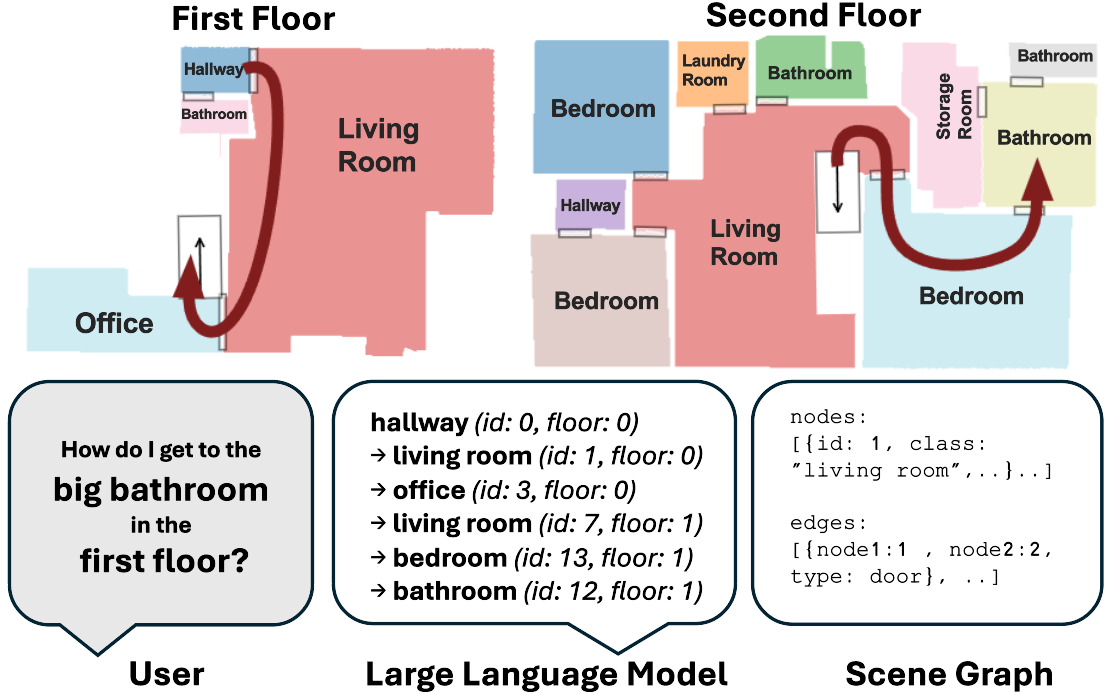}
    \caption{Navigation application based on 3D layouts and LLMs.}
    \label{fig:scene-graph}
    \vspace{-30pt}
\end{wrapfigure}
\paragraph{Analysis Experiments.}
\label{sec:analysis}

Tab.~\ref{tab:pipeline_ablation} shows the contributions in terms of F1 score of each stage in our approach.
Note that the outputs of the first and second stages (\textit{mesh} from Sec.~\ref{sec:mesh_generation} and \textit{layout skeleton} from Sec.~\ref{sec:skeleton_extraction}) are triangle meshes, which we convert to polygon sets by first applying mesh simplification (QSlim~\cite{garland1997quadricedgedecimation}), and then greedily merging adjacent triangles whose normals differ by less than 20°.
Scores drop significantly when either layout fitting or room segmentation is removed.

\paragraph{Qualitative Results.}
\label{sec:qualitative}

We show qualitative results of our approach in Fig.~\ref{fig:quali} and compare to RoomFormer~\cite{yue2023connecting} and SceneScript~\cite{avetisyan2024scenescript}.
Both baselines methods struggle with large areas consisting of multiple rooms, RoomFormer even more than SceneScript.
The baselines ere also inherently limitted to predicting rectangular primitives and cannot represent more complex shapes such as sloped ceilings \emph{(top example)}.
In Fig.~\ref{fig:loss-ablation} we visualizes qualitative results when removing loss objectives from the mesh fitting stage introduced in Sec.~\ref{sec:prototype_fitting}.

\paragraph{Applications.}
Next, we demonstrate a potential application of full-building 3D layouts.
First, we obtain the 3D scene graph where nodes represent rooms, and edges are connections between rooms (doors, stairs, etc.)
Then, we feed the scene graph in JSON format to an LLM, together with a user-prompt asking for directions.
The LLM responds with turn-by-turn directions on how to reach the desired location.
This concept is illustrated in Fig.~\ref{fig:scene-graph}.
Beyond navigation and planning, rich structural layouts also form a natural foundation for generative scene reasoning, as demonstrated by video-conditioned synthesis approaches such as VIPScene~\cite{huang2025video}, suggesting a future where large-scale layouts and video-perception models become tightly coupled.

\paragraph{Limitations.}
\label{sec:limitations}
\name{} has a longer runtime than the feed-forward baselines, taking one to two hours per \datasetname{} scene on an NVIDIA GeForce RTX 4090, compared to one to two minutes for SceneScript~\cite{avetisyan2024scenescript} and RoomFormer~\cite{yue2023connecting}.
Furthermore, \name{} occasionally struggles to remove outdoor elements perceived through large windows, which can introduce artifacts.

\section{Conclusion}
\label{sec:conclusion}
We introduced \datasetname{}, the first benchmark for evaluating 3D layout estimation in large-scale multi-floor buildings. Existing methods remain restricted to single-floor settings, and our experiments show that they struggle to parse complex buildings with multiple floors and rooms, whereas our learning-free baseline already outperforms these approaches. Looking forward, there is a clear need for learning-based models that reason across entire buildings rather than relying on heuristics that, while effective, are substantially slower than feed-forward networks.
Beyond layout estimation, the large-scale structural annotations in \datasetname{} also provide a foundation for 3D scene synthesis~\cite{huang2025video, ccelen2025housetour, bucher2025respace}, enabling the generation of coherent multi-floor environments that can in turn supply additional training data for a wide range of 3D perception models~\cite{takmaz2025search3d,vogel2024p2p, szilagyi2025slag} as in \cite{weder2024labelmaker,ji2025arkit}. In summary, we hope this benchmark will drive progress both in robust multi-floor layout estimation and in generative models that support the broader development of 3D perception methods.
\paragraph{Acknowledgments.} Francis Engelmann is supported by an SNSF PostDoc Mobility Fellowship. 
{
\small
\bibliographystyle{unsrt}
\bibliography{main}

\begin{thebibliography}{10}

\bibitem{yue2023connecting}
Yuanwen Yue, Theodora Kontogianni, Konrad Schindler, and Francis Engelmann.
\newblock {Connecting the Dots: Floorplan Reconstruction Using Two-Level
  Queries}.
\newblock In {\em International Conference on Computer Vision and Pattern
  Recognition (CVPR)}, 2023.

\bibitem{avetisyan2024scenescript}
Armen Avetisyan, Christopher Xie, Henry Howard-Jenkins, Tsun-Yi Yang, Samir
  Aroudj, Suvam Patra, Fuyang Zhang, Duncan Frost, Luke Holland, Campbell Orme,
  et~al.
\newblock {SceneScript: Reconstructing Scenes with an Autoregressive Structured
  Language Model}.
\newblock {\em European Conference on Computer Vision (ECCV)}, 2024.

\bibitem{ccelen2025housetour}
Ata {\c{C}}elen, Marc Pollefeys, Daniel Barath, and Iro Armeni.
\newblock {HouseTour: A Virtual Real Estate a (I) Gent}.
\newblock In {\em International Conference on Computer Vision (ICCV)}, 2025.

\bibitem{zheng2025wildgs}
Jianhao Zheng, Zihan Zhu, Valentin Bieri, Marc Pollefeys, Songyou Peng, and Iro
  Armeni.
\newblock {WildGS-SLAM: Monocular Gaussian Splatting SLAM in Dynamic
  Environments}.
\newblock In {\em International Conference on Computer Vision and Pattern
  Recognition (CVPR)}, 2025.

\bibitem{jin2024multiway}
Shengze Jin, Iro Armeni, Marc Pollefeys, and Daniel Barath.
\newblock {Multiway Point Cloud Mosaicking with Diffusion and Global
  Optimization}.
\newblock In {\em International Conference on Computer Vision and Pattern
  Recognition (CVPR)}, 2024.

\bibitem{lemke2024spotcompose}
Oliver Lemke, Zuria Bauer, Ren{\'e} Zurbr{\"u}gg, Marc Pollefeys, Francis
  Engelmann, and Hermann Blum.
\newblock {Spot-Compose: A Framework for Open-Vocabulary Object Retrieval and
  Drawer Manipulation in Point Clouds}.
\newblock In {\em 2nd Workshop on Mobile Manipulation and Embodied Intelligence
  at ICRA 2024}, 2024.

\bibitem{zurbrugg2024icgnet}
Ren{\'e} Zurbr{\"u}gg, Yifan Liu, Francis Engelmann, Suryansh Kumar, Marco
  Hutter, Vaishakh Patil, and Fisher Yu.
\newblock {ICGNet: A Unified Approach for Instance-Centric Grasping}.
\newblock In {\em International Conference on Robotics and Automation (ICRA)},
  2024.

\bibitem{chen2023polydiffuse}
Jiacheng Chen, Ruizhi Deng, and Yasutaka Furukawa.
\newblock {Polydiffuse: Polygonal Shape Reconstruction via Guided Set Diffusion
  Models}.
\newblock {\em International Conference on Neural Information Processing
  Systems (NeurIPS)}, 2023.

\bibitem{Structured3D}
Jia Zheng, Junfei Zhang, Jing Li, Rui Tang, Shenghua Gao, and Zihan Zhou.
\newblock {Structured3D: A Large Photo-Realistic Dataset for Structured 3D
  Modeling}.
\newblock In {\em European Conference on Computer Vision (ECCV)}, 2020.

\bibitem{avetisyan2020scenecad}
Armen Avetisyan, Tatiana Khanova, Christopher Choy, Denver Dash, Angela Dai,
  and Matthias Nießner.
\newblock {SceneCAD: Predicting Object Alignments and Layouts in RGB-D Scans}.
\newblock In {\em European Conference on Computer Vision (ECCV)}, 2020.

\bibitem{miao2024scenegraphloc}
Yang Miao, Francis Engelmann, Olga Vysotska, Federico Tombari, Marc Pollefeys,
  and D{\'a}niel~B{\'e}la Bar{\'a}th.
\newblock {Scenegraphloc: Cross-Modal Coarse Visual Localization on 3d Scene
  Graphs}.
\newblock In {\em European Conference on Computer Vision (ECCV)}, 2024.

\bibitem{wuest2025unloc}
Matthias W{\"u}est, Francis Engelmann, Ondrej Miksik, Marc Pollefeys, and
  Daniel Barath.
\newblock {UnLoc: Leveraging Depth Uncertainties for Floorplan Localization}.
\newblock {\em arXiv preprint arXiv:2509.11301}, 2025.

\bibitem{zhang2025open}
Chenyangguang Zhang, Alexandros Delitzas, Fangjinhua Wang, Ruida Zhang,
  Xiangyang Ji, Marc Pollefeys, and Francis Engelmann.
\newblock {Open-Vocabulary Functional 3d Scene Graphs for Real-World Indoor
  Spaces}.
\newblock In {\em Proceedings of the Computer Vision and Pattern Recognition
  Conference}, 2025.

\bibitem{Matterport3D}
Angel Chang, Angela Dai, Thomas Funkhouser, Maciej Halber, Matthias Niessner,
  Manolis Savva, Shuran Song, Andy Zeng, and Yinda Zhang.
\newblock {Matterport3D: Learning from RGB-D Data in Indoor Environments}.
\newblock {\em International Conference on 3d Vision (3dV)}, 2017.

\bibitem{scan2bim}
Srivathsan Murali, Pablo Speciale, Martin~R. Oswald, and Marc Pollefeys.
\newblock {Indoor Scan2BIM: Building Information Models of House Interiors}.
\newblock In {\em IEEE/RSJ International Conference on Intelligent Robots and
  Systems (IROS)}, 2017.

\bibitem{yang2019dulanetdualprojectionnetworkestimating}
Shang-Ta Yang, Fu-En Wang, Chi-Han Peng, Peter Wonka, Min Sun, and Hung-Kuo
  Chu.
\newblock {DuLa-Net: A Dual-Projection Network for Estimating Room Layouts from
  a Single RGB Panorama}.
\newblock In {\em International Conference on Computer Vision and Pattern
  Recognition (CVPR)}, 2019.

\bibitem{zou2018layoutnet}
Chuhang Zou, Alex Colburn, Qi~Shan, and Derek Hoiem.
\newblock {LayoutNet: Reconstructing the 3D Room Layout from a Single RGB
  Image}.
\newblock In {\em International Conference on Computer Vision and Pattern
  Recognition (CVPR)}, 2018.

\bibitem{liu2018floornetunifiedframeworkfloorplan}
Chen Liu, Jiaye Wu, and Yasutaka Furukawa.
\newblock {FloorNet: A Unified Framework for Floorplan Reconstruction from 3D
  Scans}.
\newblock In {\em European Conference on Computer Vision (ECCV)}, 2018.

\bibitem{OCHMANN2019251}
Sebastian Ochmann, Richard Vock, and Reinhard Klein.
\newblock {Automatic Reconstruction of Fully Volumetric 3D Building Models from
  Oriented Point Clouds}.
\newblock {\em Journal of Photogrammetry and Remote Sensing (JPRS)}, 2019.

\bibitem{cabral2014piecewiseplanarshortestpath}
Ricardo Cabral and Yasutaka Furukawa.
\newblock {Piecewise Planar and Compact Floorplan Reconstruction from Images}.
\newblock In {\em International Conference on Computer Vision and Pattern
  Recognition (CVPR)}, 2014.

\bibitem{chen2019floorspinversecadfloorplans}
Jiacheng Chen, Chen Liu, Jiaye Wu, and Yasutaka Furukawa.
\newblock {Floor-SP: Inverse CAD for Floorplans by Sequential Room-Wise
  Shortest Path}.
\newblock In {\em International Conference on Computer Vision (ICCV)}, 2019.

\bibitem{werby23hovsg}
Abdelrhman Werby, Chenguang Huang, Martin Büchner, Abhinav Valada, and Wolfram
  Burgard.
\newblock {Hierarchical Open-Vocabulary 3D Scene Graphs for Language-Grounded
  Robot Navigation}.
\newblock {\em Robotics: Science and Systems (RSS)}, 2024.

\bibitem{fedele2025superdec}
Elisabetta Fedele, Boyang Sun, Leonidas Guibas, Marc Pollefeys, and Francis
  Engelmann.
\newblock {SuperDec: 3D Scene Decomposition with Superquadric Primitives}.
\newblock In {\em International Conference on Computer Vision (ICCV)}, 2025.

\bibitem{ZInD}
Steve Cruz, Will Hutchcroft, Yuguang Li, Naji Khosravan, Ivaylo Boyadzhiev, and
  Sing~Bing Kang.
\newblock {Zillow Indoor Dataset: Annotated Floor Plans with 360º Panoramas
  and 3D Room Layouts}.
\newblock In {\em International Conference on Computer Vision and Pattern
  Recognition (CVPR)}, 2021.

\bibitem{matterportLayout3D}
Chuhang Zou, Jheng-Wei Su, Chi-Han Peng, Alex Colburn, Qi~Shan, Peter Wonka,
  Hung-Kuo Chu, and Derek Hoiem.
\newblock {Manhattan Room Layout Reconstruction from a Single 360 Image: A
  Comparative Study of State-Of-The-Art Methods}, 2020.

\bibitem{rozumnyi2023estimatinggeneric3droom}
Denys Rozumnyi, Stefan Popov, Kevis-Kokitsi Maninis, Matthias Nießner, and
  Vittorio Ferrari.
\newblock {Estimating Generic 3D Room Structures from 2D Annotations}.
\newblock In {\em International Conference on Neural Information Processing
  Systems (NeurIPS)}, 2023.

\bibitem{matterportLayout2D}
{VSIS Lab}.
\newblock {Matterport3D-Layout}, 2020.

\bibitem{dai2017scannet}
Angela Dai, Angel~X Chang, Manolis Savva, Maciej Halber, Thomas Funkhouser, and
  Matthias Nie{\ss}ner.
\newblock {Scannet: Richly-Annotated 3d Reconstructions of Indoor Scenes}.
\newblock In {\em International Conference on Computer Vision and Pattern
  Recognition (CVPR)}, 2017.

\bibitem{scasa2025pinpoint}
SCASA.
\newblock {PinPoint}.

\bibitem{turkulainen2024dnsplatter}
Matias Turkulainen, Xuqian Ren, Iaroslav Melekhov, Otto Seiskari, Esa Rahtu,
  and Juho Kannala.
\newblock {DN-Splatter: Depth and Normal Priors for Gaussian Splatting and
  Meshing}.
\newblock In {\em IEEE/CVF Winter Conference on Applications of Computer Vision
  (WACV)}, 2025.

\bibitem{schoenberger2016sfm}
Johannes~Lutz Sch\"{o}nberger and Jan-Michael Frahm.
\newblock {Structure-from-Motion Revisited}.
\newblock In {\em International Conference on Computer Vision and Pattern
  Recognition (CVPR)}, 2016.

\bibitem{kerbl3Dgaussians}
Bernhard Kerbl, Georgios Kopanas, Thomas Leimk{\"u}hler, and George Drettakis.
\newblock {3D Gaussian Splatting for Real-Time Radiance Field Rendering}.
\newblock {\em ACM Transactions On Graphics (TOG)}, 2023.

\bibitem{kazhdan2006poisson}
Michael Kazhdan, Matthew Bolitho, and Hugues Hoppe.
\newblock {Poisson Surface Reconstruction}.
\newblock In {\em Proceedings of the Fourth Eurographics Symposium on Geometry
  Processing}, 2006.

\bibitem{yin2023metric3dzeroshotmetric3d}
Wei Yin, Chi Zhang, Hao Chen, Zhipeng Cai, Gang Yu, Kaixuan Wang, Xiaozhi Chen,
  and Chunhua Shen.
\newblock {Metric3D: Towards Zero-Shot Metric 3D Prediction from a Single
  Image}.
\newblock In {\em International Conference on Computer Vision (ICCV)}, 2023.

\bibitem{jain2023oneformer}
Jitesh Jain, Jiachen Li, MangTik Chiu, Ali Hassani, Nikita Orlov, and Humphrey
  Shi.
\newblock {OneFormer: One Transformer to Rule Universal Image Segmentation}.
\newblock In {\em International Conference on Computer Vision and Pattern
  Recognition (CVPR)}, 2023.

\bibitem{lin2015microsoftcococommonobjects}
Tsung-Yi Lin, Michael Maire, Serge Belongie, Lubomir Bourdev, Ross Girshick,
  James Hays, Pietro Perona, Deva Ramanan, C.~Lawrence Zitnick, and Piotr
  Dollár.
\newblock {Microsoft COCO: Common Objects in Context}.
\newblock In {\em European Conference on Computer Vision (ECCV)}, 2014.

\bibitem{robert2024scalable}
Damien Robert, Hugo Raguet, and Loic Landrieu.
\newblock {Scalable 3D Panoptic Segmentation as Superpoint Graph Clustering}.
\newblock {\em International Conference on 3d Vision (3dV)}, 2024.

\bibitem{douglas1973algorithms}
D.~H. Douglas and T.~K. Peucker.
\newblock {Algorithms for the Reduction of the Number of Points Required to
  Represent a Digitized Line or Its Caricature}.
\newblock {\em The Canadian Cartographer}, 1973.

\bibitem{CGAL:ConstrainedDelaunayTriangulation2}
{CGAL Team}.
\newblock {\em {CGAL::Constrained\_Delaunay\_triangulation\_2}}, 2025.

\bibitem{localoutlierfactor}
Markus~M. Breunig, Hans-Peter Kriegel, Raymond~T. Ng, and J\"{o}rg Sander.
\newblock {LOF: Identifying Density-Based Local Outliers}.
\newblock In {\em Proceedings of the 2000 ACM SIGMOD International Conference
  on Management of Data}, 2000.

\bibitem{dbscan}
Martin Ester, Hans-Peter Kriegel, J{\"o}rg Sander, and Xiaowei Xu.
\newblock {A Density-Based Algorithm for Discovering Clusters in Large Spatial
  Databases with Noise}.
\newblock In {\em International Conference on Knowledge Discovery and Data
  Mining (KDD)}, 1996.

\bibitem{scannetpp}
Chandan Yeshwanth, Yueh-Cheng Liu, Matthias Nießner, and Angela Dai.
\newblock {ScanNet++: A High-Fidelity Dataset of 3D Indoor Scenes}.
\newblock In {\em International Conference on Computer Vision (ICCV)}, 2023.

\bibitem{delta_1_metric}
Lubor Ladický, Jianbo Shi, and Marc Pollefeys.
\newblock {Pulling Things Out of Perspective}.
\newblock In {\em International Conference on Computer Vision and Pattern
  Recognition (CVPR)}, 2014.

\bibitem{garland1997quadricedgedecimation}
Michael Garland and Paul~S. Heckbert.
\newblock {Surface Simplification Using Quadric Error Metrics}.
\newblock In {\em Proceedings of the 24th Annual Conference on Computer
  Graphics and Interactive Techniques}, 1997.

\bibitem{huang2025video}
Rui Huang, Guangyao Zhai, Zuria Bauer, Marc Pollefeys, Federico Tombari,
  Leonidas Guibas, Gao Huang, and Francis Engelmann.
\newblock {Video Perception Models for 3D Scene Synthesis}.
\newblock In {\em International Conference on Neural Information Processing
  Systems (NeurIPS)}, 2025.

\bibitem{bucher2025respace}
Martin~JJ Bucher and Iro Armeni.
\newblock {ReSpace: Text-Driven 3D Scene Synthesis and Editing with Preference
  Alignment}.
\newblock {\em arXiv preprint arXiv:2506.02459}, 2025.

\bibitem{takmaz2025search3d}
Ayca Takmaz, Alexandros Delitzas, Robert~W. Sumner, Francis Engelmann, Johanna
  Wald, and Federico Tombari.
\newblock {Search3D: Hierarchical Open-Vocabulary 3D Segmentation}.
\newblock {\em IEEE Robotics and Automation Letters (RA-L)}, 2025.

\bibitem{vogel2024p2p}
Mathias Vogel, Keisuke Tateno, Marc Pollefeys, Federico Tombari, Marie-Julie
  Rakotosaona, and Francis Engelmann.
\newblock {P2p-bridge: Diffusion Bridges for 3d Point Cloud Denoising}.
\newblock In {\em European Conference on Computer Vision (ECCV)}, 2024.

\bibitem{szilagyi2025slag}
Laszlo Szilagyi, Francis Engelmann, and Jeannette Bohg.
\newblock {SLAG: Scalable Language-Augmented Gaussian Splatting}.
\newblock {\em IEEE Robotics and Automation Letters (RA-L)}, 2025.

\bibitem{weder2024labelmaker}
Silvan Weder, Hermann Blum, Francis Engelmann, and Marc Pollefeys.
\newblock {LabelMaker: Automatic Semantic Label Generation from RGB-D
  Trajectories}.
\newblock In {\em International Conference on 3d Vision (3dV)}, 2024.

\bibitem{ji2025arkit}
Guangda Ji, Silvan Weder, Francis Engelmann, Marc Pollefeys, and Hermann Blum.
\newblock {Arkit Labelmaker: A New Scale for Indoor 3d Scene Understanding}.
\newblock In {\em Proceedings of the Computer Vision and Pattern Recognition
  Conference}, 2025.

\end{thebibliography}
}

\end{document}